\documentclass{article}
\usepackage{ijcai22}

\usepackage{times}
\usepackage{soul}
\usepackage[hyphens]{url}
\usepackage[hidelinks]{hyperref}
\usepackage[utf8]{inputenc}
\usepackage[small]{caption}
\usepackage{graphicx}
\usepackage{amsmath}
\usepackage{amsthm}
\usepackage{amssymb}
\usepackage{booktabs}
\usepackage{algorithm}
\usepackage{algorithmic}
\usepackage{multibib}
\newcites{SM}{Supplementary References}
\newcommand{\std}[1]{\emph{$\pm$ #1}}

\title{Monitoring Vegetation From Space at Extremely Fine Resolutions via Coarsely-Supervised Smooth U-Net}

\author{
Joshua Fan$^1$\footnote{Contact Author}\and
Di Chen$^1$\and
Jiaming Wen$^{2}$\and
Ying Sun$^{2}$\And
Carla P. Gomes$^1$\\
\affiliations
$^1$Department of Computer Science, Cornell University\\
$^2$School of Integrative Plant Science -- Soil and Crop Sciences Section, Cornell University\\
\emails
jyf6@cornell.edu, di@cs.cornell.edu, jw2495@cornell.edu,  ys776@cornell.edu, gomes@cs.cornell.edu
}
\begin{document}

\maketitle

\begin{abstract}
Monitoring vegetation productivity at extremely fine resolutions is valuable for real-world agricultural applications, such as detecting crop stress and providing early warning of food insecurity. Solar-Induced Chlorophyll Fluorescence (SIF) provides a promising way to directly measure plant productivity from space. However, satellite SIF observations are only available at a coarse spatial resolution, making it impossible to monitor how individual crop types or farms are doing. 
This poses a challenging \emph{coarsely-supervised regression} (or \emph{downscaling}) task; at training time, we only have SIF labels at a coarse resolution (3km), but we want to predict SIF at much finer spatial resolutions (e.g. 30m, a 100x increase). We also have additional fine-resolution input features, but the relationship between these features and SIF is unknown. To address this, we propose \emph{Coarsely-Supervised Smooth U-Net (CS-SUNet)}, a novel method for this coarse supervision setting. 
CS-SUNet combines the expressive power of deep convolutional networks with novel regularization methods based on prior knowledge (such as a smoothness loss) that are crucial for preventing overfitting. Experiments show that CS-SUNet resolves fine-grained variations in SIF more accurately than existing methods.

\end{abstract}

\section{Introduction}

\noindent Monitoring crop growth is vital for preventing food insecurity and supporting humanitarian efforts to address the UN Sustainable Development Goal 2 (``Zero Hunger'') \cite{peng2020assessing}.\footnote{Appendices can be found at \url{http://joshuafan.github.io/files/SIF_supp.pdf}}
This is particularly crucial due to the impacts of climate change on agriculture \cite{ortiz2021anthropogenic}. Thus, there have been many attempts to observe vegetation progress from space via remote sensing. A common approach is to use vegetation indices that can be computed from satellite imagery, such as NDVI. However, these indices are only indicative of vegetation ``greenness'', and cannot directly measure photosynthetic activity \cite{li2019global}. On the other hand, Solar-Induced Chlorophyll Fluoresence (SIF) holds the potential to directly monitor plant productivity, as it has mechanistic linkages to plant photosynthesis, and provides more accurate and timely information about plant growth \cite{kira2020extraction}. Satellite-based Solar-Induced Chlorophyll Fluorescence (SIF) can outperform traditional vegetation indices for crop yield prediction \cite{peng2020assessing}, as well as for monitoring the effects of drought and heat stress on vegetation productivity \cite{song2020satellite}. 


However, despite recent progress, SIF continues to be a very difficult signal to measure from space \cite{duveiller2020spatially}. 
Most satellite SIF measurements are noisy, and are thus only available at a coarse spatial resolution, such as 3-5km \cite{yu2019high}. 
Within a large region of many square kilometers, SIF can vary dramatically, depending on crop type, management practices, and genetic varieties.  For example, the soybean fields in a particular region may be growing very well, while the corn fields may be experiencing stress. Yet existing satellite measurements cannot resolve this important variation \cite{kira2020extraction}. 

 
There have been several attempts to predict SIF at finer spatial resolutions, given coarse-resolution SIF measurements and additional fine-resolution variables \cite{yu2019high,duveiller2020spatially,wen2020framework}.  
This task is sometimes known as ``statistical downscaling.'' These works train a supervised machine learning model at a coarse scale, to map from auxiliary input variables (averaged across an entire coarse tile) to the SIF of that tile. Then this trained model is applied at a fine scale, using the fine-resolution input variables to predict fine-resolution SIF. Since the input variables are averaged across a large region during training, information about fine-grained spatial variation is lost. Thus, these works only attempt to resolve SIF down to a 0.05-degree ($\approx 5$ km) resolution. They do not attempt to resolve SIF at super-fine resolutions such as 30 meters; our results show that these methods are inaccurate at such fine resolutions. 




This problem poses a challenging \emph{coarsely-supervised regression} task. We want to learn a model that maps from fine-resolution input features to fine-resolution SIF, yet the only SIF labels available at training time are at a much coarser resolution. We note that this task (inferring a fine-resolution map of a variable, given only coarse-resolution observations and other fine-resolution auxiliary data) is not exclusive to SIF, but is applicable to many problems in diverse fields, such as improving the resolution of climate model outputs \cite{wood2004hydrologic}, soil moisture \cite{peng2017review}, and malaria risk \cite{sturrock2014fine}.

To address this issue, we propose a novel technique, \emph{Coarsely-Supervised Smooth U-Net (CS-SUNet)}, which can produce SIF predictions at a 30-meter resolution, even though it is only trained with SIF labels at a 3km resolution (100x coarser). CS-SUNet takes in a fine-resolution input image, and outputs a SIF prediction for each pixel; the average of the pixel predictions is trained to match the true coarse SIF for the entire 3km image.  The U-Net architecture is good at producing localized predictions, as the prediction for a pixel primarily depends on nearby pixels' features \cite{wang2020weakly}. We further enhance the architecture to make the localization more precise and reduce instability. Still, the coarse-resolution SIF labels are an extremely weak form of supervision, and do not provide enough constraints for the pixel predictions. However, we show that CS-SUNet can still learn to predict accurately through regularization techniques inspired by prior knowledge.

\paragraph{Our contributions.} \textbf{(1)} We propose CS-SUNet, a novel approach to coarsely-supervised regression problems, which builds on the expressive power and localization abilities of U-Net.
To our knowledge, this type of approach has never been applied to resolving fine-resolution SIF or similar ``coarsely-supervised regression'' problems with spatial structure. \textbf{(2)} We find that naively training a U-Net can result in a unique form of overfitting (by outputting extreme pixel predictions that happen to average to the right value). Thus, to regularize the model, we design a smoothness loss that encourages pixels with similar features to map to similar SIF. We also use early stopping based on a small fine-resolution validation set.  \textbf{(3)} We evaluate our method on real-world remote sensing and SIF data. CS-SUNet predicts SIF at fine resolutions more accurately than all baselines, reducing RMSE by almost 10\%.

\section{Related Work}

\paragraph{SIF Downscaling: Mechanistic Approaches.} There have been several attempts to downscale coarse-resolution SIF measurements. One line of work uses a mechanistic Light Use Efficiency modeling approach to improve the spatial resolution of the GOME-2 satellite's SIF measurements from 0.5-degree to 0.05-degree \cite{duveiller2020spatially}. Hand-crafted theoretical models are fitted to establish semi-empirical relationships between SIF and explanatory variables (vegetation indices, evapotranspiration, and temperature). These methods impose strong assumptions on the nature of these relationships, and assume that they hold across all land cover types.

\paragraph{SIF Downscaling: Data-Driven Approaches.} There have also been attempts to use completely data-driven \emph{statistical downscaling} methods. These studies use standard machine learning algorithms, such as Artificial Neural Network \cite{yu2019high} or regression trees~\cite{li2019global}, to learn an empirical relationship between averaged coarse-resolution input features (e.g. reflectance) and coarse-resolution SIF. This relationship is then applied on fine-resolution reflectance to estimate fine-resolution SIF. These approaches work well at a relatively coarse resolution (such as downscaling from 0.5 to 0.05 degrees), but become less accurate at finer resolutions, since coarse-resolution SIF labels used in training typically represent the SIF of a large region with multiple land cover types. However, we are interested in predicting SIF for individual fields of a single crop type. Because reflectance data for the entire large tile is averaged together, the model cannot observe variation between crop types within a tile during training, or learn crop-specific patterns.
Recently,~\cite{kira2020extraction} addressed this ``SIF unmixing'' problem by estimating crop-specfic SIF for corn and soybean portions of a larger region. They produce a training dataset of artificial ``mixed pixels'', which were generated by mixing pure-corn and pure-soy tiles together in random proportions. Then they train a neural network to estimate how much of the SIF was from the corn areas vs. the soybean areas. However, this approach requires that pure tiles for each crop type exist, which is not the case for most crops other than corn and soybean. 
\paragraph{Statistical Downscaling in Other Fields.} The ``statistical downscaling'' approach has also been applied to numerous problems in geoscience and other fields, including downscaling soil moisture \cite{peng2017review} and other remotely sensed variables. 
As previously mentioned, these methods discard information about spatial variability within the large regions while training. Bias-corrected spatial disaggregation \cite{wood2004hydrologic} is another method to downscale projections from climate models, but it requires extensive ground-truth fine-resolution labels. Hierarchical Bayesian models have also been used for downscaling, such as for species distributions and malaria risk~\cite{sturrock2014fine}. 
However, they model fine-resolution probabilities as a simple linear function of the inputs, which limits the expressiveness of the learned functions. Some works have proposed methods based on Gaussian Processes ~\cite{tanaka2019refining}, but they are computationally intractable on large datasets (such as our setting where we want to predict for millions of pixels).
 
\paragraph{Multiple-Instance Learning.} In the machine learning literature, there are some prior works on multiple-instance learning, where each training label is over a group of examples. While most of these works focus on classification, a few papers consider regression tasks (\emph{aggregate output regression}). \cite{musicant2007supervised} presented adaptations of supervised learning algorithms to this setting. \cite{kotzias2015group} studied this in the context of fine-grained sentiment analysis of online reviews, and proposed adding a smoothness loss term, which is similar to the idea in our paper. \cite{zhang2020learning} proposed probabilistic models for this setting, but the method they derive for regression tasks is just a simple L2 loss between the predicted and true aggregated outputs, similar to the coarse loss term in our approach. Unlike these works, our approach utilizes the spatial structure in our data via U-Net.

\paragraph{Supervised Single-Image Super-Resolution.} Many papers have studied \emph{single-image super-resolution} using convolutional neural networks (CNNs), where the goal is to learn a function mapping directly from a coarse-resolution image to the fine-resolution version. For example, \cite{liu2020climate} use a novel fusion CNN architecture to downscale climate model variables to finer resolution. However, these methods require abundant fine-resolution data for training, which is not available for SIF. By contrast, CS-SUNet does not use fine-resolution labels at training time, so it is usable in settings where fine-resolution labels are scarce.


\paragraph{Image Segmentation.} CS-SUNet is inspired by techniques originally developed for image segmentation, specifically the U-Net architecture~\cite{unet}. The U-Net produces a prediction for each pixel in the original image, based on the contents of that pixel and neighboring pixels. The U-Net is typically used for classification, but CS-SUNet is adapted for regression.

\paragraph{Weakly-Supervised Pixel Predictions.} While image segmentation tasks are usually trained using pixel-level labels, some works use weaker forms of supervision (e.g. image-level labels). Recently, \cite{wang2020weakly} showed that the U-Net can accurately segment remote sensing imagery (e.g. classifying each pixel as cropland or not), even if it is only trained with image-level labels. This is largely due to the U-Net's impressive localization abilities, which we harness in our approach.
~\cite{wang2018diabetic} show that CNNs can provide localization abilities for regression tasks also. They train a CNN to predict diabetic retinopathy from a retina image, and propose a method, Regression Activation Mapping, to determine which regions of the image have the greatest influence on the final prediction. By contrast, our paper explicitly predicts the label (SIF) for every pixel in the image.






\section{Datasets}

We summarize the datasets used in this paper. More details (including citations) can be found in Appendices E and F.





\subsection{Input Features}

\paragraph{Landsat Surface Reflectance.} Satellite imagery is available from Landsat at a fine 30m resolution, every 16 days. For each pixel, there are 7 bands (channels). We zero out pixels with missing data and create a mask to mark this.


\paragraph{FLDAS land data.} Temperature, rainfall, and surface radiation data are available from FLDAS (Famine Early Warning Systems Network (FEWS NET) Land Data Assimilation System), at a $10$km resolution. We linearly interpolate these variables to a 30m resolution to match the other variables.

\paragraph{Land cover data.} Land cover data is available from the Cropland Data Layer at 30m resolution.  Each pixel is labelled with a land cover type. For each land cover type that makes up more than 1\% of the dataset, we create a binary mask of which pixels are covered by that type.


\subsection{SIF Labels}


To evaluate model performance, we need some ground-truth fine-resolution SIF data. The Chlorophyll Fluorescence Imaging Spectrometer (CFIS) provides SIF observations at fine resolutions for extremely limited areas. We aggregate the CFIS observations to a 30m resolution, matching the Landsat grid. Then, we produce datasets at different resolutions: $\{30, 90, 150, 300, 600\}$ meters for evaluation, and 3km for training (to simulate the realistic setting in which only coarse-resolution SIF observations are available).




OCO-2~\cite{sun2017oco} provides additional SIF labels at a coarse spatial resolution, which we can also use for coarse-resolution training. The original footprints cover areas of around $1.3 \times 2.25$ km at nadir. We grid these to a 3km resolution (on the same grid as our CFIS 3km dataset). To correct between the differing wavelengths of the different instruments, we multiply OCO-2 SIF values by 1.11 to match CFIS. (This scaling factor was empirically determined by fitting a model on OCO-2, evaluating it on CFIS, and determining what scaling factor produced the optimal fit.) 

Since SIF observations are noisy, we filter our dataset to only include pixels and tiles with enough observations (so that noise can be reduced through averaging). We also remove tiles with many missing features. See Appendix G for details.

\subsection{Study Region and Time Range}

\begin{figure}
\centering
\includegraphics[width=0.9\columnwidth]{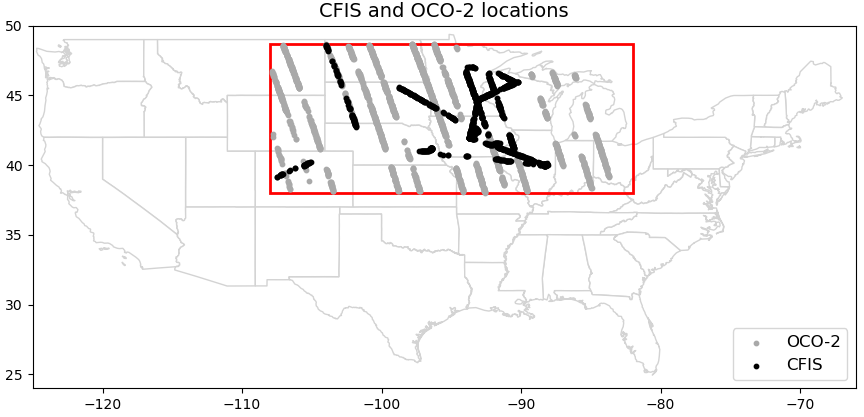}
\caption{Locations of CFIS (black) and OCO-2 (dark gray) tiles in our study region (red box).}
\label{map}
\end{figure}

All tiles in our dataset correspond to a coarse-resolution CFIS or OCO-2 SIF label that lies within the Midwest US, from 38 to 48.7 degrees N and 108 to 82 degrees W. The region is plotted in Figure \ref{map}. As shown in the figure, even coarse-resolution SIF datasets are sparse geographically, so machine learning is needed to fill in the gaps and predict at a higher resolution. We extract input feature data (reflectance, crop cover maps, etc.) from the same time periods as the CFIS SIF observations: June 15-29, 2016 and August 1-16, 2016. We also use OCO-2 observations from those time periods. 

\begin{figure*}
\centering
\includegraphics[width=0.9\textwidth]{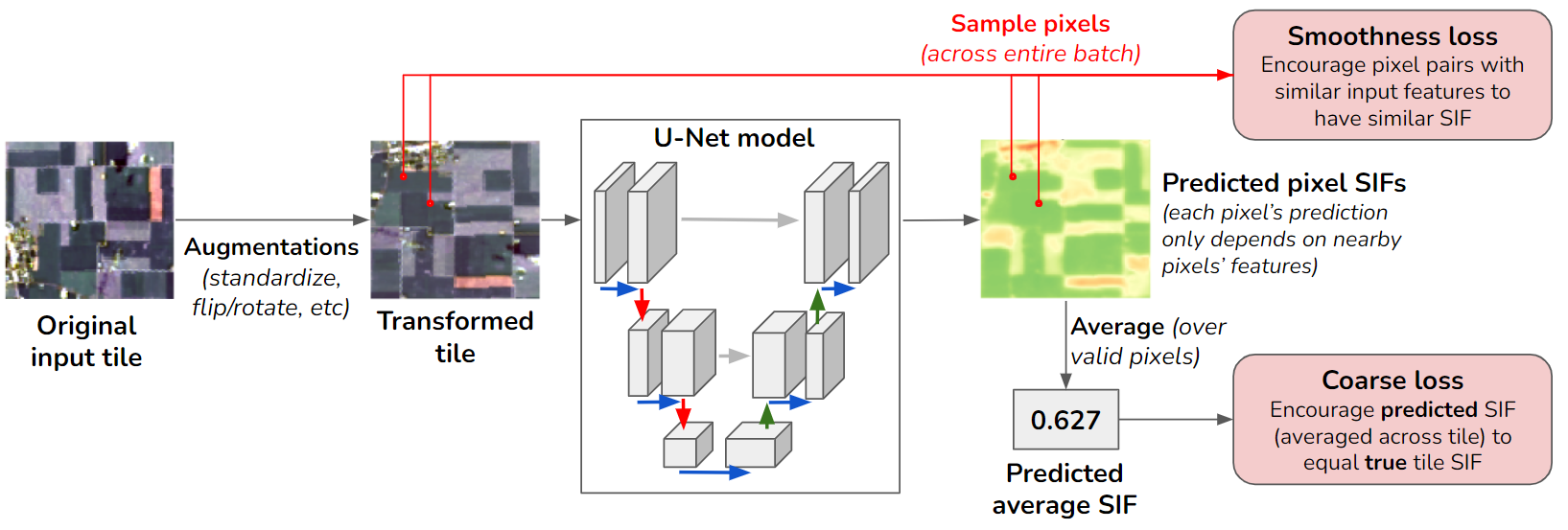}
\caption{Coarsely-Supervised Smooth U-Net (CS-SUNet). At each iteration, we first apply some augmentations to the original input tile (satellite images, etc). Then we pass it through the U-Net to obtain pixel SIF predictions. Finally, we compute the coarsely-supervised loss and the smoothness regularization loss, and backpropagate to improve the model.}
\label{unet}
\end{figure*}

\subsection{Data Preprocessing}

We use a total of 22 input features: 7 Landsat reflectance bands (plus 1 binary mask indicating missing Landsat data), 3 FLDAS variables (temperature, rainfall, surface radiation), and 11 land cover binary masks. All features were resampled to a fine resolution of 30 meters. For each coarse (3km) SIF label, we extracted the corresponding tile of feature data. These tiles are 3-dimensional tensors, of size $22$ (features)$ \times 100 \times 100$ (pixels). For the continuous variables, we standardized them to have mean 0 and variance 1, and clipped them to $[-3, 3]$ standard deviations from the mean.


We randomly assign each tile to one of 3 sets: train (60\%), validation (20\%), and test (20\%); we ensure that CFIS and OCO-2 tiles that overlap end up in the same set. The model is trained on coarse-resolution (tile-level) SIF labels from the train tiles; there are a total of 712 CFIS and 1390 OCO-2 tiles in the train set. Then, we select hyperparameters based on NRMSE on the \emph{fine-resolution CFIS validation} set.

\section{Methods}
\label{sec:methods}

We propose CS-SUNet, a novel approach for coarsely-supervised regression tasks. CS-SUNet utilizes the localization abilities of the U-Net architecture, adapts it to our coarsely-supervised regression setting, and employs regularization techniques informed by prior knowledge, such as a smoothness loss and early stopping.

\subsection{U-Net Architecture}

CS-SUNet makes use of the U-Net architecture~\cite{unet}, which was  originally designed for segmenting biomedical images using relatively little training data. The U-Net is a fully convolutional neural network which can take in input images of any size. It consists of a contracting path which aggregates information from local regions (left side), and an expanding path which combines broader large-area information with precise local characteristics (right side). The final output is a map, with a prediction for each pixel.


In our implementation, we used 2 upsampling and 2 downsampling blocks, with $\{64, 128\}$ hidden units. For half of the convolutional layers, we decreased the filter size from $3$ to $1$ to ensure that each pixel's prediction only depends on nearby pixels' features. The output is a single scalar per pixel. We removed the batch normalization layers as they caused training to be unstable; they do not appear to be a good fit for regression tasks like ours that depend on the absolute magnitude of the inputs (see Appendix G). Further model details can be found in Appendix J.

\subsection{Coarsely Supervised Training}

CS-SUNet is trained in a coarsely-supervised fashion. Our training set consists of pairs $\{(\mathbf{X}_i,  y_i)\}_{i=1}^N$: $\mathbf{X}_i \in \mathbb{R}^{C \times H \times W}$ is a large tile ``image'' with $C$ channels, height $H$, and width $W$, and $y_i \in \mathbb{R}$ is the ground-truth average (coarse) SIF for tile $i$.


We train a U-Net model $f_{\theta}$ that takes in large tile $\mathbf{X}_i$ and predicts SIF for each pixel, $\mathbf{Z}_i \in \mathbb{R}^{H \times W}$. Let $z_i^{(p)} \in \mathbb{R}$ be the predicted SIF of pixel $p$. During training, we do not have pixel-level SIF labels. Instead, we train the model by optimizing a combination of two loss functions: (1) a coarse loss that encourages the average predicted SIF of the entire tile to equal the true SIF of the tile, and (2) a smoothness loss that encourages pixel pairs with similar input features to map to similar SIF predictions.

The coarse loss is defined as the Mean Squared Error between the true and predicted coarse tile SIFs; this is similar to~\cite{zhang2020learning}. We average predictions over $P$, the set of valid fine pixels that contributed to the tile's coarse SIF:
$$\mathcal{L}_{coarse}(y_i, \mathbf{Z}_i) = \left(y_i - \frac{1}{|P|} \sum_{p \in P} z_{i}^{(p)} \right)^2$$
However, achieving a low coarse loss does not guarantee that the fine-resolution predictions will be accurate. It is possible for the U-Net to output extreme pixel predictions that still produce the correct coarse-tile average. We find that this occurs frequently in practice (see Appendix B). To reduce this unique form of overfitting, we introduce a smoothness loss \cite{kotzias2015group}, which encourages pixels with similar input features to map to similar SIFs. To compute this loss, for each batch we randomly sample a set of pixels $S$; in our experiments we sample 1000 pixels per batch. For each pair of pixels in $S$, we compute a similarity score between their input feature vectors ($\mathbf{x}^{(j)}$, $\mathbf{x}^{(k)}$), and multiply that by the squared difference between their predicted SIFs ($z^{(j)}$, $z^{(k)}$):
$$\mathcal{L}_{smooth}(\mathbf{X}, \mathbf{Z}) = \sum_{j \in S} \sum_{k \in S} \text{sim}(\mathbf{x}^{(j)}, \mathbf{x}^{(k)}) \cdot (z^{(j)} - z^{(k)})^2$$
This places a large penalty if two pixels with high input similarity have very different SIF, as this represents overfitting. The similarity function can be customized based on one's prior knowledge of the problem. We set the similarity to 0 if pixels $j$ and $k$ are of different land cover types. Otherwise, if they are of the same land cover type, we define their input similarity as
$$\text{sim}(\mathbf{x}^{(j)}, \mathbf{x}^{(k)}) = \exp \left(-\tau \|\mathbf{x}_{ref}^{(j)} - \mathbf{x}_{ref}^{(k)} \|_2^2 \right)$$
Here $\mathbf{x}_{ref}^{(j)}$ is the reflectance feature vector for pixel $j$. Note that if $\tau$ is positive, the similarities are bounded between 0 and 1. $\tau$ is a hyperparameter that controls how quickly the similarities decay towards 0; we set this to $0.5$. Appendix C discusses the impact of the smoothness loss, and Appendix D motivates the design of this loss.

The final loss is a weighted combination of the coarse loss and the smoothness loss. For a batch,
$$\mathcal{L}(\mathbf{X}, \mathbf{Z}, \mathbf{y}) = \frac{1}{B} \left[ \sum_{i=1}^B \mathcal{L}_{coarse}(y_i, \mathbf{Z}_i) \right] + \lambda \mathcal{L}_{smooth}(\mathbf{X}, \mathbf{Z})$$
where $\lambda$ is a hyperparameter that balances the loss terms. We optimize this loss via backpropagation. Figure \ref{unet} depicts the training process.

\subsection{Early Stopping}



CS-SUNet does not look at any fine-resolution SIF data during training. However, during validation, CS-SUNet does peek at the model's performance on the \emph{fine-resolution validation set} to select which epoch's model to use. Like the smoothness loss, early stopping encourages ``input smoothness'' (pixels with similar input features should map to similar SIF) \cite{rosca2020case}, which is crucial for regularizing our model. The fine-resolution data is only used to determine how much regularization is necessary, so it may not need to be very similar to the evaluation setting. If we do not have any fine-resolution data at all, it is difficult to determine the optimal amount of regularization, but our smoothness loss can still prevent overfitting on later epochs (see Appendix C).

\section{Experiments} \label{results}

We evaluate our pixel-level SIF predictions against airborne measurements from CFIS, which are available at a 30-meter resolution in limited regions. To reduce measurement noise, we only evaluate on pixels for which CFIS had at least 30 observations and where the SIF value was at least 0.1. We use Normalized Root Mean Squared Error (NRMSE) and $R^2$ as evaluation metrics. (NRMSE is Root Mean Squared Error, normalized by the average SIF of the training dataset.)

\subsection{Training Details}

The CS-SUNet approach is compared against existing ``statistical downscaling'' baselines that have been used in papers such as~\cite{yu2019high,li2019global}. These involve averaging each input feature (channel) over all valid pixels in the tile, and training a model to predict SIF from the feature averages. We consider 4 types of models: Ridge Regression, Gradient Boosting Regressor, Random Forest, and a fully-connected artificial neural network (ANN).

We also include a few novel baselines. The ``Pixel NN'' approach involves feeding each pixel through the same neural network, averaging all pixels' predictions across a tile, and then training using coarse loss (tile-level supervision); this substitutes the U-Net architecture with an per-pixel architecture where each pixel's prediction is only influenced by that pixel's features. 
The ``vanilla U-Net'' approach involves training a U-Net with only the coarse loss, but without regularization (early stopping or smoothness loss). 
Finally, ``U-Net fine supervision'' trains a U-Net using fine-resolution labels, and represents an upper limit on our model performance if we were actually able to train on fine-resolution labels.

We train all deep models with the AdamW optimizer \cite{adamw}. For each method, a grid-search was performed over hyperparameter values (see Appendix I for details); we selected the hyperparameters that performed best on the \emph{fine-resolution} validation set. For CS-SUNet, we used a batch size of 128, learning rate 3e-4, weight decay 1e-4, $\tau=0.5$, and $\lambda=0.5$. For methods that involve randomness, we took the average of 3 runs using different random seeds. For tiles where the coarse-resolution SIF label is observed during training, we also include a trivial ``predict coarse'' baseline where we predict the SIF of every pixel to be the same as the SIF of the entire tile.  


\subsection{Results}


\begin{table}
\centering
\begin{tabular}{lcc}
\toprule
\textbf{Method} & \textbf{NRMSE} ($\downarrow$)& \textbf{$R^2$} ($\uparrow$) \\ \midrule
\emph{Predict coarse} & \emph{0.248} & \emph{0.372} \\ 
Ridge Regression & 0.203 & 0.582 \\
Gradient Boosting & 0.212 & 0.541 \\
Random Forest & 0.220 \std{0.002} & 0.507 \std{0.008} \\
ANN & 0.215 \std{0.005} & 0.528 \std{0.020} \\
Pixel NN & 0.208 \std{0.004} & 0.561 \std{0.016} \\
Vanilla U-Net & 0.223 \std{0.005} & 0.492 \std{0.021} \\
CS-SUNet & \textbf{0.184} \std{0.005} & \textbf{0.656} \std{0.019} \\
\bottomrule
\end{tabular}
\caption{Results on interpolating CFIS to 30m resolution, on train tiles where only \emph{coarse-resolution} SIF was known. The ``predict coarse'' row is predicting the coarse SIF for each pixel. ``ANN'' is artificial neural network. For methods that involve randomness, we report the average $\pm$ the standard deviation over 3 runs.}
\label{interpolate}
\end{table}

\begin{table}
\centering
\begin{tabular}{lcc} \toprule
\textbf{Method} & \textbf{NRMSE} ($\downarrow$) & \textbf{$R^2$} ($\uparrow$)\\ \midrule
Ridge Regression & 0.205 & 0.361 \\
Gradient Boosting & 0.216 & 0.289 \\ 
Random Forest & 0.229 \std{0.001} & 0.199 \std{0.006} \\
ANN & 0.215 \std{0.002} & 0.300 \std{0.016} \\
Pixel NN & 0.207 \std{0.002} & 0.348 \std{0.013} \\
Vanilla U-Net & 0.241 \std{0.011} & 0.113 \std{0.081}  \\
CS-SUNet & \textbf{0.185} \std{0.002} & \textbf{0.481} \std{0.013} \\
\emph{U-Net fine supervision} & \emph{0.161 \std{0.003}} & \emph{0.605 \std{0.014}} \\ \bottomrule
\end{tabular}
  \caption{Results on extrapolating CFIS to 30m resolution, on test tiles where even coarse-resolution SIF is unknown.}
  \label{extrapolate}
\end{table}

\begin{figure*}[bth]
\centering
\includegraphics[width=0.75\textwidth]{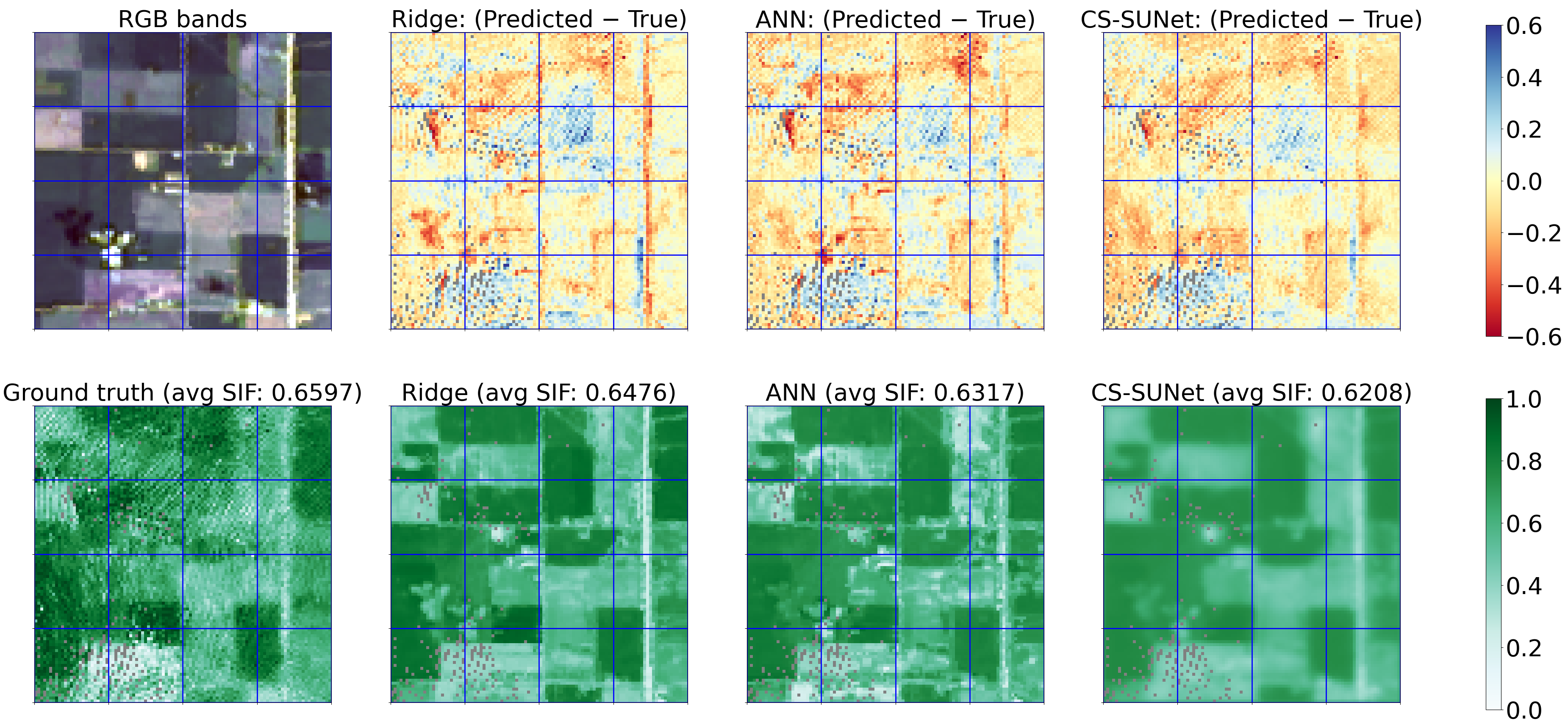} 
\caption{Example result. \textbf{Top-left:} RGB bands, used as input features. \textbf{Bottom-left:} ground-truth CFIS SIF (gray pixels are missing data). \textbf{Other bottom images:} predictions with different methods (CS-SUNet is rightmost). \textbf{Other top images:} error maps (yellow is accurate, blue is over-prediction, red is under-prediction). 
CS-SUNet is generally more accurate than other methods -- for example, there are fewer dark red or blue pixels (which indicate major errors).}
\label{example_result}
\end{figure*}

Table \ref{interpolate} presents the results for fine pixels in train tiles (i.e. the coarse-resolution SIF of the entire tile was known during training, but not the fine-resolution SIF of the individual pixels). The model is tasked with interpolating the fine-resolution SIF from the coarse-resolution labels. Since we know the coarse-resolution SIF of these tiles, we multiply each tile's pixel predictions by a per-tile constant, such that the average of the pixel predictions exactly equals the true coarse SIF. Table \ref{extrapolate} presents results for fine pixels where not even coarse-resolution SIF was seen during training; it tests the model's ability to generalize to ``gaps'' where SIF observations were not directly available. (Note that the relatively lower $R^2$ in Table \ref{extrapolate} is because there is less total variation in these tiles; the absolute error magnitudes (NRMSE) are similar to Table \ref{interpolate}.)



According to both metrics, CS-SUNet  outperforms standard ``statistical downscaling'' methods in both settings, reducing NRMSE by 9-10\% and increasing $R^2$ by 13-33\% (relative) over the best baseline, Ridge Regression. This improvement occurs across multiple land cover types and resolutions (see Appendix A). CS-SUNet outperforms the ``Pixel NN'' architecture, which shows that the prediction accuracy for a pixel can be improved if you look at neighboring pixels' features, not just that pixel's features. Notably, the ``Vanilla U-Net'' actually does poorly -- simply applying a U-Net out of the box does not work, due to the ``overfitting to coarse labels'' problem described earlier. In order for the U-Net to work, it needs to be highly regularized, which we achieve in this paper by early stopping and a novel smoothness loss.

While an NRMSE of 18-19\% is not perfect, note that even a model trained with fine-resolution labels can only achieve an NRMSE of 16\% on the test tiles, so our performance is not far off. In any case, increasing the resolution of SIF measurements by 100x is an extremely challenging task, and our method provides significant improvements.

\paragraph{Qualitative evaluation.} Figure \ref{example_result} presents an example of the predictions outputted by various measurements. All methods provide useful results; even the basic Ridge Regression method is able to identify fields with higher and lower SIF. However, 
 the error maps on the top row show that 
 CS-SUNet produces more reliable predictions, as there are fewer dark blue or red pixels (which indicate large errors). In addition, CS-SUNet takes the spatial context of pixels into consideration, so the predictions outputted by CS-SUNet are much more smooth and consistent across local regions.





\paragraph{Limitations.} The evaluation of our models is limited by the fact that there is significant noise in both SIF measurements and Landsat reflectance (and there may be a few days' gap between them). Our approach for SIF prediction requires Landsat reflectance data, yet Landsat data is unavailable or noisy in some places/times due to cloud cover. This could be alleviated in the future as more satellites are deployed.

\section{Conclusion}

We present Coarsely-Supervised Smooth U-Net (CS-SUNet), which is capable of  predicting SIF at a very fine resolution (30m), even when only coarse-resolution (3km) SIF measurements are available. Due to its localization properties, CS-SUNet is able to figure out which farms within a large tile had higher and lower SIF. CS-SUNet harnesses the expressive power of deep convolutional networks, yet avoids overfitting due to its smoothness loss and early stopping.



Although CS-SUNet's predictions are not perfect, they clearly outperform existing state-of-the-art methods, and can provide valuable vegetation information. Even noisy fine-resolution SIF estimates may facilitate improvements in crop yield prediction and monitoring, as SIF contains detailed information about plant photosynthesis that is not captured in vegetation indices such as NDVI \cite{peng2020assessing}, which are simple combinations of a few spectral bands.
Moreover, in addition to SIF, CS-SUNet could also potentially be applied to make fine-resolution predictions of any numerical variable that is only available at coarse resolution, if there exists additional fine-resolution data that is correlated with the variable of interest. Such applications could include predicting soil moisture, precipitation, disease prevalence, and species distributions at fine resolutions.


\clearpage

\section*{Acknowledgements}

This research was supported by NSF award CCF-1522054
(Expeditions in Computing) and the NSF NRT Digital Plant Science Fellowship (1922551). We thank Junwen Bai for proofreading and providing valuable suggestions.

\bibliographystyle{named}
\bibliography{ref}

\begin{thebibliography}{}

\bibitem[\protect\citeauthoryear{Bannari \bgroup \em et al.\egroup
  }{1995}]{vegetationindices}
A~Bannari, D~Morin, F~Bonn, and AjRsr Huete.
\newblock A review of vegetation indices.
\newblock {\em Remote sensing reviews}, 13(1-2):95--120, 1995.

\bibitem[\protect\citeauthoryear{Bjorck \bgroup \em et al.\egroup
  }{2018}]{bjorck2018understanding}
Nils Bjorck, Carla~P Gomes, Bart Selman, and Kilian~Q Weinberger.
\newblock Understanding batch normalization.
\newblock {\em Advances in neural information processing systems}, 31, 2018.

\bibitem[\protect\citeauthoryear{Frankenberg \bgroup \em et al.\egroup
  }{2018}]{CFIS}
Christian Frankenberg, Philipp K{\"o}hler, Troy~S Magney, Sven Geier, Peter
  Lawson, Mark Schwochert, James McDuffie, Darren~T Drewry, Ryan Pavlick, and
  Andreas Kuhnert.
\newblock The chlorophyll fluorescence imaging spectrometer (cfis), mapping far
  red fluorescence from aircraft.
\newblock {\em Remote sensing of environment}, 217:523--536, 2018.

\bibitem[\protect\citeauthoryear{Masek \bgroup \em et al.\egroup
  }{2006}]{Landsat}
Jeffrey~G Masek, Eric~F Vermote, Nazmi~E Saleous, Robert Wolfe, Forrest~G Hall,
  Karl~Fred Huemmrich, Feng Gao, Jonathan Kutler, and Teng-Kui Lim.
\newblock A landsat surface reflectance dataset for north america, 1990-2000.
\newblock {\em IEEE Geoscience and Remote Sensing Letters}, 3(1):68--72, 2006.

\bibitem[\protect\citeauthoryear{McNally \bgroup \em et al.\egroup
  }{2017}]{fldas}
Amy McNally, Kristi Arsenault, Sujay Kumar, Shraddhanand Shukla, Pete Peterson,
  Shugong Wang, Chris Funk, Christa~D Peters-Lidard, and James~P Verdin.
\newblock A land data assimilation system for sub-saharan africa food and water
  security applications.
\newblock {\em Scientific data}, 4(1):1--19, 2017.

\bibitem[\protect\citeauthoryear{NASS}{2016}]{cdl}
USDA NASS.
\newblock Usda national agricultural statistics service cropland data layer.
\newblock {\em USDA-NASS, Washington, DC}, 2016.

\bibitem[\protect\citeauthoryear{Sun \bgroup \em et al.\egroup
  }{2017}]{sun2017oco}
Ying Sun, Christian Frankenberg, Jeffery~D Wood, DS~Schimel, Martin Jung, Luis
  Guanter, DT~Drewry, Manish Verma, Albert Porcar-Castell, Timothy~J Griffis,
  et~al.
\newblock Oco-2 advances photosynthesis observation from space via
  solar-induced chlorophyll fluorescence.
\newblock {\em Science}, 358(6360):eaam5747, 2017.

\end{thebibliography}


\begin{thebibliography}{}

\bibitem[\protect\citeauthoryear{Duveiller \bgroup \em et al.\egroup
  }{2020}]{duveiller2020spatially}
G.~Duveiller, F.~Filipponi, S.~Walther, P.~K{\"o}hler, C.~Frankenberg,
  L.~Guanter, and A.~Cescatti.
\newblock A spatially downscaled sun-induced fluorescence global product for
  enhanced monitoring of vegetation productivity.
\newblock {\em Earth System Science Data}, 12(2):1101--1116, 2020.

\bibitem[\protect\citeauthoryear{Kira and Sun}{2020}]{kira2020extraction}
O.~Kira and Y.~Sun.
\newblock Extraction of sub-pixel c3/c4 emissions of solar-induced chlorophyll
  fluorescence (sif) using artificial neural network.
\newblock {\em ISPRS Journal of Photogrammetry and Remote Sensing},
  161:135--146, 2020.

\bibitem[\protect\citeauthoryear{Kotzias \bgroup \em et al.\egroup
  }{2015}]{kotzias2015group}
D.~Kotzias, M.~Denil, N.~De~Freitas, and P.~Smyth.
\newblock From group to individual labels using deep features.
\newblock In {\em Proceedings of the 21th ACM SIGKDD international conference
  on knowledge discovery and data mining}, pages 597--606, 2015.

\bibitem[\protect\citeauthoryear{Li and Xiao}{2019}]{li2019global}
X.~Li and J.~Xiao.
\newblock A global, 0.05-degree product of solar-induced chlorophyll
  fluorescence derived from oco-2, modis, and reanalysis data.
\newblock {\em Remote Sensing}, 11(5):517, 2019.

\bibitem[\protect\citeauthoryear{Liu \bgroup \em et al.\egroup
  }{2020}]{liu2020climate}
Y.~Liu, A.~R. Ganguly, and J.~Dy.
\newblock Climate downscaling using ynet: A deep convolutional network with
  skip connections and fusion.
\newblock In {\em Proceedings of the 26th ACM SIGKDD International Conference
  on Knowledge Discovery \& Data Mining}, pages 3145--3153, 2020.

\bibitem[\protect\citeauthoryear{Loshchilov and Hutter}{2018}]{adamw}
I.~Loshchilov and F.~Hutter.
\newblock Decoupled weight decay regularization.
\newblock In {\em International Conference on Learning Representations}, 2018.

\bibitem[\protect\citeauthoryear{Musicant \bgroup \em et al.\egroup
  }{2007}]{musicant2007supervised}
D.~R. Musicant, J.~M. Christensen, and J.~F. Olson.
\newblock Supervised learning by training on aggregate outputs.
\newblock In {\em Seventh IEEE International Conference on Data Mining (ICDM
  2007)}, pages 252--261. IEEE, 2007.

\bibitem[\protect\citeauthoryear{Ortiz-Bobea \bgroup \em et al.\egroup
  }{2021}]{ortiz2021anthropogenic}
A.~Ortiz-Bobea, T.~R. Ault, C.~M. Carrillo, R.~G. Chambers, and D.~B. Lobell.
\newblock Anthropogenic climate change has slowed global agricultural
  productivity growth.
\newblock {\em Nature Climate Change}, 11(4):306--312, 2021.

\bibitem[\protect\citeauthoryear{Peng \bgroup \em et al.\egroup
  }{2017}]{peng2017review}
J.~Peng, A.~Loew, O.~Merlin, and N.~E.~C. Verhoest.
\newblock A review of spatial downscaling of satellite remotely sensed soil
  moisture.
\newblock {\em Reviews of Geophysics}, 55(2):341--366, 2017.

\bibitem[\protect\citeauthoryear{Peng \bgroup \em et al.\egroup
  }{2020}]{peng2020assessing}
B.~Peng, K.~Guan, W.~Zhou, C.~Jiang, C.~Frankenberg, Y.~Sun, L.~He, and
  P.~K{\"o}hler.
\newblock Assessing the benefit of satellite-based solar-induced chlorophyll
  fluorescence in crop yield prediction.
\newblock {\em International Journal of Applied Earth Observation and
  Geoinformation}, 90:102126, 2020.

\bibitem[\protect\citeauthoryear{Ronneberger \bgroup \em et al.\egroup
  }{2015}]{unet}
O.~Ronneberger, P.~Fischer, and T.~Brox.
\newblock U-net: Convolutional networks for biomedical image segmentation.
\newblock In {\em International Conference on Medical image computing and
  computer-assisted intervention}, pages 234--241. Springer, 2015.

\bibitem[\protect\citeauthoryear{Rosca \bgroup \em et al.\egroup
  }{2020}]{rosca2020case}
M.~Rosca, T.~Weber, A.~Gretton, and S.~Mohamed.
\newblock A case for new neural networks smoothness constraints.
\newblock In {\em ''I Can't Believe It's Not Better!''NeurIPS 2020 workshop},
  2020.

\bibitem[\protect\citeauthoryear{Song \bgroup \em et al.\egroup
  }{2020}]{song2020satellite}
Y.~Song, J.~Wang, and L.~Wang.
\newblock Satellite solar-induced chlorophyll fluorescence reveals heat stress
  impacts on wheat yield in india.
\newblock {\em Remote Sensing}, 12(20):3277, 2020.

\bibitem[\protect\citeauthoryear{Sturrock \bgroup \em et al.\egroup
  }{2014}]{sturrock2014fine}
H.~J.~W. Sturrock, J.~M. Cohen, P.~Keil, A.~J. Tatem, A.~Le~Menach, N.~E.
  Ntshalintshali, M.~S. Hsiang, and R.~D. Gosling.
\newblock Fine-scale malaria risk mapping from routine aggregated case data.
\newblock {\em Malaria journal}, 13(1):1--9, 2014.

\bibitem[\protect\citeauthoryear{Sun \bgroup \em et al.\egroup
  }{2017}]{sun2017oco}
Y.~Sun, C.~Frankenberg, J.~D. Wood, D.~S. Schimel, M.~Jung, L.~Guanter, D.~T.
  Drewry, M.~Verma, A.~Porcar-Castell, T.~J. Griffis, et~al.
\newblock Oco-2 advances photosynthesis observation from space via
  solar-induced chlorophyll fluorescence.
\newblock {\em Science}, 358(6360):eaam5747, 2017.

\bibitem[\protect\citeauthoryear{Tanaka \bgroup \em et al.\egroup
  }{2019}]{tanaka2019refining}
Y.~Tanaka, T.~Iwata, T.~Tanaka, T.~Kurashima, M.~Okawa, and H.~Toda.
\newblock Refining coarse-grained spatial data using auxiliary spatial data
  sets with various granularities.
\newblock In {\em Proceedings of the AAAI Conference on Artificial
  Intelligence}, volume~33, pages 5091--5099, 2019.

\bibitem[\protect\citeauthoryear{Wang and Yang}{2018}]{wang2018diabetic}
Z.~Wang and J.~Yang.
\newblock Diabetic retinopathy detection via deep convolutional networks for
  discriminative localization and visual explanation.
\newblock In {\em Workshops at the thirty-second AAAI conference on artificial
  intelligence}, 2018.

\bibitem[\protect\citeauthoryear{Wang \bgroup \em et al.\egroup
  }{2020}]{wang2020weakly}
S.~Wang, W.~Chen, S.~M. Xie, G.~Azzari, and D.~B. Lobell.
\newblock Weakly supervised deep learning for segmentation of remote sensing
  imagery.
\newblock {\em Remote Sensing}, 12(2):207, 2020.

\bibitem[\protect\citeauthoryear{Wen \bgroup \em et al.\egroup
  }{2020}]{wen2020framework}
J.~Wen, P.~K{\"o}hler, G.~Duveiller, N.~C. Parazoo, T.~S. Magney, G.~Hooker,
  L.~Yu, C.~Y. Chang, and Y.~Sun.
\newblock A framework for harmonizing multiple satellite instruments to
  generate a long-term global high spatial-resolution solar-induced chlorophyll
  fluorescence (sif).
\newblock {\em Remote Sensing of Environment}, 239:111644, 2020.

\bibitem[\protect\citeauthoryear{Wood \bgroup \em et al.\egroup
  }{2004}]{wood2004hydrologic}
A.~W. Wood, L.~R. Leung, V.~Sridhar, and D.~P. Lettenmaier.
\newblock Hydrologic implications of dynamical and statistical approaches to
  downscaling climate model outputs.
\newblock {\em Climatic change}, 62(1):189--216, 2004.

\bibitem[\protect\citeauthoryear{Yu \bgroup \em et al.\egroup
  }{2019}]{yu2019high}
L.~Yu, J.~Wen, C.~Y. Chang, C.~Frankenberg, and Y.~Sun.
\newblock High-resolution global contiguous sif of oco-2.
\newblock {\em Geophysical Research Letters}, 46(3):1449--1458, 2019.

\bibitem[\protect\citeauthoryear{Zhang \bgroup \em et al.\egroup
  }{2020}]{zhang2020learning}
Y.~Zhang, N.~Charoenphakdee, Z.~Wu, and M.~Sugiyama.
\newblock Learning from aggregate observations.
\newblock {\em Advances in Neural Information Processing Systems},
  33:7993--8005, 2020.

\end{thebibliography}


\appendix

\section*{Appendices for \emph{Monitoring Vegetation From Space at Extremely Fine Resolutions via Coarsely-Supervised Smooth U-Net}}

\section{Additional Results} \label{app:additional_results}

We break down the results on ``train tiles'' by land cover type (Table \ref{croptypes}) and resolution (Table \ref{resolution}). CS-SUNet performs best across all land cover types and resolutions, indicating the robustness of our approach.

\begin{table*}
 \centering
\begin{tabular}{lcccc} \toprule
\textbf{Method} & \textbf{Grassland} & \textbf{Corn} & \textbf{Soybean} & \textbf{Deciduous Forest} \\ \midrule
\emph{Trivial: predict coarse} & \emph{0.255} & \emph{0.225} & \emph{0.278}  & \emph{0.196} \\ 
Ridge Regression &
0.230 & 0.183 & 0.212 & 0.192  \\ 
Gradient Boosting &
0.223 & 0.198 & 0.219 & 0.206 \\ 
Random Forest & 0.237 \std{0.002} & 0.212 \std{0.001} & 0.221 \std{0.003} & 0.211 \std{0.003} \\
ANN &
0.247 \std{0.005} & 0.193 \std{0.009} & 0.226 \std{0.007} & 0.193 \std{0.006} \\ 
Pixel NN & 0.219 \std{0.002} & 0.189 \std{0.006} & 0.217 \std{0.004} & 0.207 \std{0.004} \\
Vanilla U-Net & 0.211 \std{0.010} & 0.208 \std{0.008} & 0.258 \std{0.011} & 0.179 \std{0.002}\\
CS-SUNet & \textbf{0.188} \std{0.007} & \textbf{0.167} \std{0.003} & \textbf{0.204} \std{0.008} & \textbf{0.171} \std{0.003} \\ \bottomrule
\end{tabular}
  \caption{NRMSE by land cover type, 30m pixels in train tiles (where 3km coarse-resolution labels were seen during training). Lower is better.}
  \label{croptypes}
\end{table*}

\begin{table*}
  \centering
\begin{tabular}{lccccc} \toprule
\textbf{Method} & \textbf{30m} & \textbf{90m} & \textbf{150m} & \textbf{300m} & \textbf{600m} \\ \midrule
\emph{Trivial: predict coarse} & \emph{0.248} & \emph{0.245} & \emph{0.229} & \emph{0.203} & \emph{0.165} \\ 
Ridge Regression &
0.203 & 0.177 & 0.155 & 0.127 & 0.096 \\ 
Gradient Boosting &
0.212 & 0.183 & 0.163 & 0.134 & 0.105 \\ 
Random Forest & 0.220 \std{0.002} & 0.184 \std{0.001} & 0.163 \std{0.000} & 0.135 \std{0.000} & 0.111 \std{0.000}\\
ANN &
0.215 \std{0.005} & 0.186 \std{0.002} & 0.164 \std{0.002} & 0.133 \std{0.002} & 0.108 \std{0.006} \\
Pixel NN & 0.208 \std{0.004} & 0.179 \std{0.001} & 0.158 \std{0.001}  & 0.130 \std{0.001} & 0.099 \std{0.002} \\
Vanilla U-Net & 0.223 \std{0.005} & 0.194 \std{0.001} & 0.176 \std{0.002} & 0.150 \std{0.002} & 0.121 \std{0.003}
 \\
CS-SUNet & \textbf{0.184} \std{0.005} & \textbf{0.168} \std{0.002} & \textbf{0.150} \std{0.001} & \textbf{0.123} \std{0.000} & \textbf{0.094} \std{0.001}
\\ \bottomrule
\end{tabular}
  \caption{NRMSE by resolution, train tiles (where 3km coarse-resolution labels were seen during training). Lower is better.}
  \label{resolution}
\end{table*}

\begin{figure}[h]
\centering
\includegraphics[width=0.9\columnwidth]{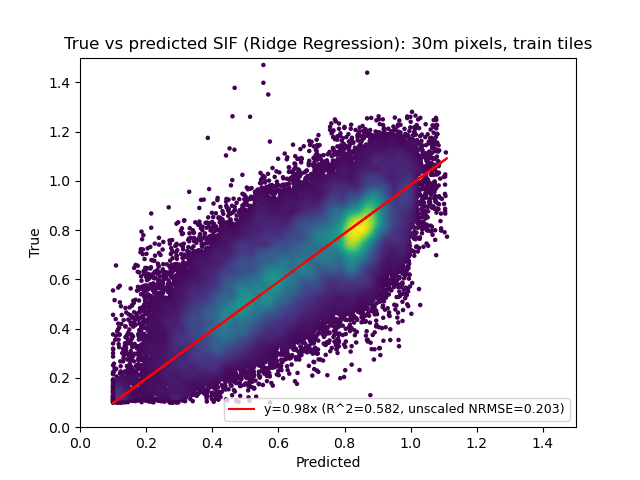}
\includegraphics[width=0.9\columnwidth]{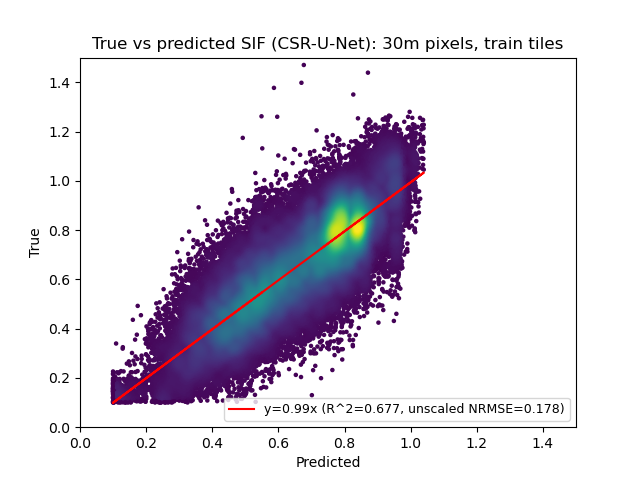} 
\caption{Ground-truth vs. predicted SIF for 30m pixels, on train tiles. \textbf{Top:} Ridge Regression, \textbf{Bottom:} CS-SUNet}.
\label{scatter}
\end{figure}

Figure \ref{scatter} presents scatterplots of true vs. predicted SIF for fine-resolution (30m) pixels, for CS-SUNet and the best performing baseline (Ridge Regression). CS-SUNet's predictions are generally closer to the identity function, indicating more accurate predictions.

\section{Importance of Regularization} \label{app:regularization}

\begin{figure}
\centering
\includegraphics[width=0.95\columnwidth]{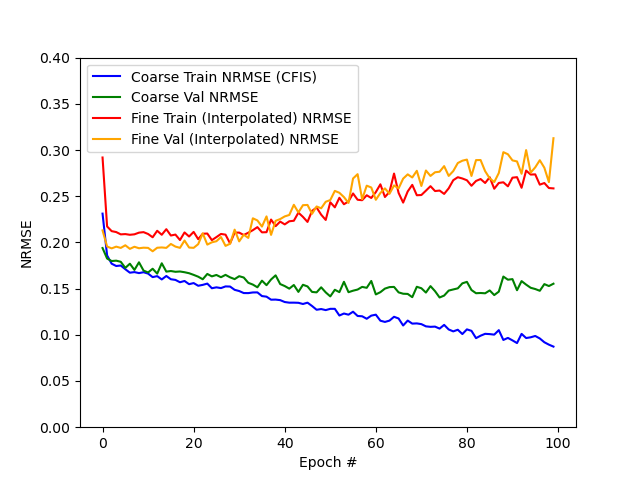}
\caption{Losses over time in a model that is not sufficiently regularized. Note that the coarse-resolution losses (blue/green) continue decreasing for while, but the fine-resolution losses (red/orange) go up quickly after epoch 20-30, indicating overfitting to the coarse labels.}
\label{loss}
\end{figure}

As described in the paper, we find that early stopping (based on a small fine-resolution validation set) and/or using the smoothness loss is critical to producing reasonable results. Without regularization, the model can overfit in a way that is unique to coarsely-supervised regression tasks. For example, Figure \ref{loss} plots losses over time for a model run that is not sufficiently regularized. Note that the coarse-resolution validation loss (green) decreases continuously until around epoch 60, but the fine-resolution losses (red/orange) start increasing after epoch 20-30. In later epochs, the model is making pixel predictions that produce the correct tile average SIF, but are inaccurate for the individual pixels.

\begin{figure}[bth]
\centering
\includegraphics[width=0.45\columnwidth]{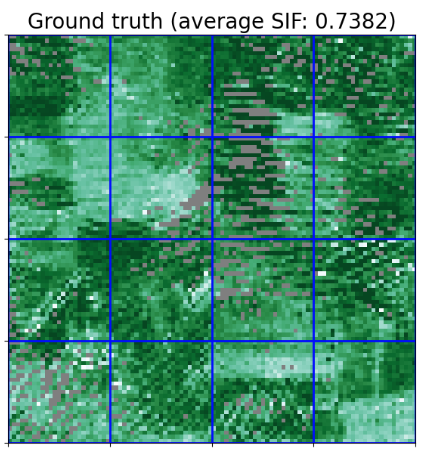} 
\includegraphics[width=0.45\columnwidth]{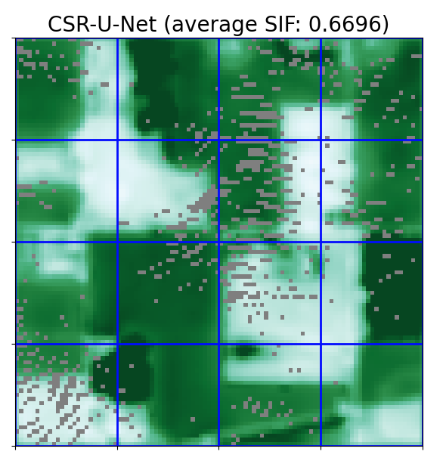} 
\caption{Example of overfitting. \textbf{Left:} ground-truth SIF map. \textbf{Right:} prediction by a U-Net that is over-trained. Note that the average tile SIFs are not too different, but the model tends to output extremely low and high values that do not reflect reality.}
\label{overfitting}
\end{figure}

Specifically, we observed that over-trained models tend to output extreme maps; the model learns how to keep pushing the predictions for low SIF regions down and high SIF regions up, in a way that maintains the correct average. An example of this is shown in Figure \ref{overfitting}.

 Given the coarse nature of training labels, the only way for the model to avoid this is to look across multiple tiles, and ensure that pixels with similar features in different tiles have similar SIF predictions. In other words, the model needs to be well-regularized. So far, we found that early stopping (based on a fine-resolution validation set) and a smoothness loss are the most effective forms of regularization for our problem. Early stopping is known to implicitly encourage smooth models where similar inputs map to similar outputs \cite{rosca2020case}.
 
 \section{Impact of Smoothness Loss} \label{app:smoothness}
 
\begin{figure}[bth]
\centering
\includegraphics[width=0.99\columnwidth]{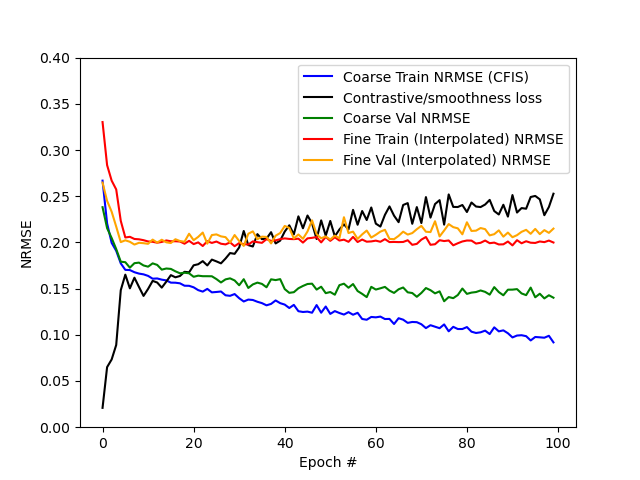}
\caption{Losses over time with smoothness loss. Note that the red/orange lines are a lot flatter in later epochs, indicating that the fine-resolution performance does not degrade as much due to overfitting.}
\label{smoothnesslosscurves}
\end{figure}

 While early stopping already does a lot to regularize the model, we find that incorporating a smoothness loss can further prevent the model from overfitting in later epochs, as it explicitly penalizes overfitted models (it places a large penalty if pixels with similar input features map to very different SIF). As shown in Figure \ref{smoothnesslosscurves}, having a smoothness loss term tends to stabilize the fine-resolution losses later in the training process, and prevents them from getting much worse.
 
We then investigate varying the strength of the smoothness loss $\lambda$ (fixing $\tau=0.5$); the results are shown in Table \ref{lambda}. When $\lambda=0$ (no smoothness loss), the performance is already quite good, likely because early stopping is enough to yield a well-regularized model. Increasing $\lambda$ can slightly improve performance on the fine-resolution validation set, and we find that the model performs well over a wide range of $\lambda$ up to 2. When $\lambda$ gets too high, model performance degrades -- having too much weight on the smoothness loss encourages the model to output similar predictions everywhere.

\begin{table}
 \centering
\begin{tabular}{lc} \toprule
\textbf{$\lambda$} & \textbf{NRMSE} \\
& (fine val pixels) \\ \midrule
0 & 0.185  \\
0.01 & 0.186 \\
0.1 & 0.184 \\
0.3 & 0.182 \\
0.5 & \textbf{0.182} \\
0.7 & 0.182 \\
1 & 0.182 \\
2 & 0.187 \\
5 & 0.197 \\
10 & 0.208 \\
100 & 0.261 \\ \bottomrule
\end{tabular}
  \caption{Impact of changing $\lambda$ (smoothness loss weight)}
  \label{lambda}
\end{table}

Finally, we check how robust our model is to changes in $\tau$, the ``spread parameter'' in the smoothness kernel (fixing $\lambda=0.5$). The results are shown in Table \ref{tau}. The model performs well over a wide range of $\tau$, from 0.1 to 100. A high value of $\tau$ means that the similarity function decays towards 0 quickly, so it is rare for pixels to be considered similar; the effect is similar to removing the smoothness loss.
\begin{table}
 \centering
\begin{tabular}{lc} \toprule
\textbf{$\tau$} & \textbf{NRMSE} \\
& (fine val pixels) \\ \midrule
0.01 & 0.213 \\
0.1 & 0.187 \\
0.2 & 0.183 \\
0.3 & 0.182 \\
0.5 & \textbf{0.182} \\
0.7 & 0.182 \\
1 & 0.183 \\
10 & 0.185 \\
100 & 0.186 \\ \bottomrule
\end{tabular}
  \caption{Impact of changing $\tau$ (spread parameter in smoothness loss)}
  \label{tau}
\end{table}

\section{Rationale Behind Smoothness Loss} \label{app:rationale_smoothness}

\begin{figure}[bth]
\centering
\includegraphics[width=0.95\linewidth]{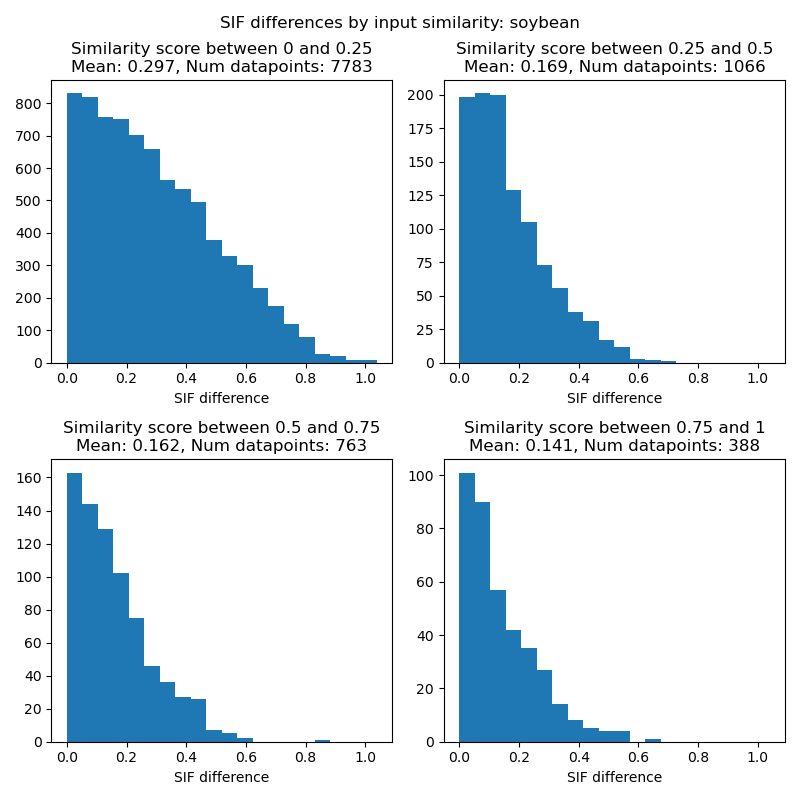}
\caption{Distributions of SIF differences, between pixels of varying similarity levels}
\label{sifdifferences}
\end{figure}

To motivate the design of the smoothness loss, we sample random pairs of pixels of each land cover type. At different ranges of input (reflectance) similarity, we show the distribution of SIF differences in Figure \ref{sifdifferences}. These plots are for soybean, but the trend holds generally.

For pixels with low input similarity (top left), SIF varies a lot. But for pixels with high input siimlarity (bottom right), the differences in SIF are much smaller. This demonstrates that pixels that are similar in input features are also likely to have similar SIFs. Thus, for pixels with similar input features, the model should be penalized if it outputs SIF predictions that are too different -- this indicates overfitting.



\section{Dataset Summary} \label{app:dataset}

Table \ref{input_features} summarizes the input features used, and Table \ref{sif_datasets} summarizes the SIF datasets used. For SIF data, we use the CFIS dataset \citeSM{CFIS} for fine-resolution measurements (as well as aggregated coarse-resolution measurements), and the OCO-2 dataset \citeSM{sun2017oco} for additional coarse-resolution measurements.

\begin{table}[h]
\fontsize{9}{10}\selectfont
  \centering
\begin{tabular}{lc} \toprule
\textbf{Dataset} & \textbf{\# vars} \\ \midrule
\textbf{Landsat surface reflectance} & 8 \\
Blue, green, red, near infrared, etc. & \\ \midrule
\textbf{FLDAS land data} & 3 \\ 
Rainfall, temperature, radiation & \\ \midrule
\textbf{CDL land cover types} (binary masks) & 11 \\
Corn, soybean, grassland, forest, etc. & \\ \bottomrule
\end{tabular}
  \caption{Summary of input features used}
  \label{input_features}
\end{table}

\begin{table}[h]
\centering
\begin{tabular}{ccc} \toprule
\textbf{Dataset} & \textbf{Resolution} & \textbf{\# 3km labels in train set} \\ \midrule
\textbf{CFIS} & 30 m & 712  \\
\textbf{OCO-2} & 3 km & 1390 \\ \bottomrule
\end{tabular}
  \caption{Summary of SIF datasets used}
  \label{sif_datasets}
\end{table}

\section{List of Features} \label{app:features}

Here is a full list of input features used in our tiles. We downloaded Landsat and Cropland Data Layer features from Google Earth Engine, and the FLDAS features from NASA Earth Data portal.\footnote{\url{https://disc.gsfc.nasa.gov/datasets/FLDAS_NOAH01_C_GL_M_001/summary?keywords=\%22MERRA-2\%22}} Each 30m pixel has a value for each of these features. \\

\textbf{Landsat surface reflectance} \citeSM{Landsat}:
\begin{enumerate}
    \item Ultra blue surface reflectance (435-451 nm)
    \item Blue surface reflectance (452-512 nm)
    \item Green surface reflectance (533-590 nm)
    \item Red surface reflectance (636-673 nm)
    \item Near infrared surface reflectance (851-879 nm)
    \item Shortwave infrared 1 surface reflectance (1566-1651 nm)
    \item Shortwave infrared 2 surface reflectance (2107-2294 nm)
    \item Missing reflectance mask (1 if reflectance data is missing, e.g. due to cloud cover)
\end{enumerate}

\textbf{FLDAS land data features} \citeSM{fldas}:
\begin{enumerate}
    \item Rainfall flux (kg m-2 s-1)
    \item Surface downward shortwave radiation (W m-2)
    \item Surface air temperature (K)
\end{enumerate}

\textbf{Cropland Data Layer land cover types} \citeSM{cdl}:
\begin{enumerate}
    \item Grassland/pasture
    \item Corn
    \item Soybean
    \item Deciduous Forest
    \item Evergreen Forest
    \item Developed/Open Space
    \item Woody Wetlands
    \item Open Water
    \item Alfalfa
    \item Developed/Low Intensity
    \item Developed/High Intensity
\end{enumerate}

\section{Data Filtering} \label{app:filtering}


We filter our dataset to exclude pixels and tiles with insufficient or noisy data. For CFIS SIF labels, geographic coverage is extremely limited (as the measurements were taken from an airplane), so most tiles only have SIF labels for some pixels. As a geographic coverage requirement, we only include $3\times 3$ km tiles that contain at least $1000$ pixels with at least 1 CFIS measurement. Our models are trained to predict the average SIF over the pixels with CFIS data, rather than over all pixels within the tile.

At evaluation time, we compare our algorithms' fine-resolution SIF predictions with the ground-truth CFIS SIF labels at different resolutions, including $\{30, 90, 150, 300, 600\}$ meters. To reduce measurement noise in the fine-resolution SIF labels, we only evaluate on small pixels that have at least $30$ soundings (observations) and have SIF $> 0.1$ (because low SIF values are difficult to measure accurately). In the next section, we show the impact of varying the number of soundings; the more soundings a pixel has, the more reliable the SIF label should be, since random noise is reduced through averaging.  At resolutions of greater than 30 meters, we also require that $90\%$ of the 30m pixels within the larger pixel have at least 1 CFIS measurement.

For OCO-2 SIF, we remove tiles that have less than 3 soundings, and tiles with SIF below $0.1$.

Also, we remove tiles where more than 50\% Landsat pixels are missing or unreliable (as defined by the Landsat QA band), since in this case there is not enough data to accurately predict the entire-tile SIF. (Also, Landsat pixels that are near cloudy areas tend to be noisy.) We removed tiles where less than 50\% of the tile is covered by one of the common land cover types we use (that make up more than 1\% of our dataset).

\section{Impact of data quality (number of soundings)} \label{app:soundings}

As SIF observations are noisy and contain a lot of measurement error, we analyze the impact of this by comparing performance against the minimum number of soundings per pixel. If a pixel has more soundings (observations), its noise should be reduced through averaging (due to the Central Limit Theorem). The results are shown in Table \ref{soundings}.

\begin{table}
 \centering
\begin{tabular}{lcc} \toprule
\textbf{Soundings} & \textbf{Ridge NRMSE} & \textbf{CS-SUNet NRMSE} \\
& (fine train pixels) & (fine train pixels) \\ \midrule
1 & 0.261 & 0.242  \\
5 & 0.248 & 0.230 \\
10 & 0.241 & 0.223 \\
20 & 0.212 & 0.190 \\
30 & 0.203 & 0.178 \\ \bottomrule
\end{tabular}
  \caption{Impact of changing the minimum number of soundings.}
  \label{soundings}
\end{table}

As we increase the number of soundings and thus reduce measurement error, performance steadily improves, indicating that data noise is a significant issue in evaluation. We mitigate this by only evaluating on pixels with at least 30 soundings to reduce noise in the labels, but our results may still be impacted by data noise.

\section{Training Details} \label{app:training}

We train on one NVIDIA Tesla V100 GPU with 16GB memory, on the Linux CentOS 7 operating system. For CS-SUNet, training a model for 100 epochs takes roughly 45 minutes. We used the following libraries with Python 3.7: Matplotlib 3.3.4, Numpy 1.18.1, Pandas 1.1.3, PyTorch 1.7.0, Scikit-Learn 0.24.1, Scipy 1.4.1. 
For all methods that involve randomness, we report the average and standard deviation using three random seeds: \{0, 1, 2\}.

For the baseline methods, we did a grid search over hyperparameters, and chose the configuration that performed best on the fine-resolution validation set. For Ridge Regression, we selected the regularization parameter $\alpha$ from $\{0.01, 0.1, 1, 10, 100, 1000, 10000\}$; we chose $\alpha = 100$. For Gradient Boosting Regressor, we selected the maximum number of iterations from $\{100, 300, 1000\}$, and the maximum depth of the tree from $\{2, 3, None\}$. We chose 100 iterations and max depth of 2. For Random Forest, we selected the number of trees from $\{10, 30, 100, 300, 1000\}$ and the max features per split from $\{2, 5, None\}$; we selected 300 trees, and set max features to 5. For the fully-connected artificial neural network, we selected hidden layer sizes from $\{(100), (20, 20), (100, 100), (100, 100, 100)\}$,  initial learning rate from $\{10^{-2}, 10^{-3}, 10^{-4}\}$, and set the maximum number of iterations to 10,000. We chose hidden layer sizes of $(100, 100, 100)$ and an initial learning rate of $0.001$.

For CS-SUNet, Pixel NN, and Vanilla U-Net methods, we used the AdamW optimizer and a batch size of 128. 
Then we did a hyperparameter search, considering learning rates from \{1e-4, 3e-4, 1e-3\} and weight decay from \{0, 1e-4, 1e-3\}. We only tuned based on random seed 0. For Pixel NN, as well as ``U-Net fine supervision'', we chose learning rate 1e-3 and weight decay 1e-3. For the Vanilla U-Net, we chose learning rate 1e-4 and weight decay 0. (Note that for the Vanilla U-Net, we used the model at epoch 100; we did not use early stopping.) For CS-SUNet, we chose learning rate 3e-4 and weight decay 1e-4. 

For CS-SUNet's smoothness loss, we considered values of $\tau$ (spread) from $\{0.01, 0.1, 0.2, 0.3, 0.5, 0.7, 1, 10, 100\}$ and $\lambda$ (weight) from $\{0, 0.01, 0.1, 0.3, 0.5, 0.7, 1, 2, 5, 10, 100\}$. We found that $\tau = 0.5$ and $\lambda=0.5$ gave the best result out of the combinations we tried, although there were many similarly good options. (Coincidentally, setting $\tau$ and $\lambda$ to be equal seems to work well.)

In terms of model architecture, we used a smaller version of U-Net with 2 downsampling and 2 upsampling blocks, with \{64, 128, 256\} hidden units. We start with a 1x1 convolution for a pixel encoder, followed by a rectified linear unit (ReLU), and then 2 downsampling blocks. Each downsampling block consists of the following sequence: ($2 \times 2$ average pooling, convolutional layer with filter size 3, ReLU, convolutional layer with filter size 1, ReLU). Note that reducing the second convolutional layer's filter size to 1 reduces the receptive field of each pixel and ensures better localization.

We use 2 upsampling blocks; each involves upsampling the feature map, and then the sequence: (convolutional layer with filter size 3, ReLU, convolutional layer with filter size 1, ReLU). Finally, we concatenate the output feature map with the feature map from the higher-resolution layer in the contracting path. 

\section{Effect of batch normalization} \label{app:batchnorm}

Overall, we found that batch normalization actually made results slightly worse, and made training more unstable. To confirm this, we tried multiple learning rates with and without batch normalization. We used a larger batch size of 256 to make the batch statistics more stable. Even still, removing batch normalization improved performance. Batch normalization does allow us to use higher learning rates \citeSM{bjorck2018understanding}, but this does not improve results.
\begin{table}[h]
\centering
\begin{tabular}{ccc} \toprule
\textbf{Learning rate} & \textbf{No batch norm} & \textbf{With batch norm} \\
& (fine val NRMSE) & (fine val NRMSE) \\ \midrule
1e-4 & \textbf{0.183} & 0.204 \\
1e-3 & 0.183 & 0.193 \\ 
1e-2 & 0.194 & 0.204 \\
1e-1 & 0.310 & 0.204 \\ \bottomrule
\end{tabular}
  \caption{Effect of batch normalization}
  \label{batchnorm}
\end{table}

We hypothesize that this is because our problem actually depends on the absolute intensity values of the input images. Batch normalization introduces significant noise by scaling by the mean and standard deviation \emph{of each batch}, which removes information about the raw intensities. This phenomenon is also reported in a blog post.\footnote{\url{https://towardsdatascience.com/pitfalls-with-dropout-and-batchnorm-in-regression-problems-39e02ce08e4d}} This is fine for tasks with natural images, which tend to be more invariant to shifts in light intensity and color. However, when monitoring vegetation, the absolute intensities matter; for example, traditional vegetation indices are simple mathematical formulas based on the absolute intensity values of different channels in these remote sensing images  \citeSM{vegetationindices}.

\section{Data and Code Availability}

We intend to submit an expanded version of this paper to a journal, such as in the field of remote sensing. After the journal article is published, we plan to make the entire dataset and codebase publicly available. All of the raw data used already comes from publicly available sources.

\section{Evaluation Metrics}

We evaluate our model on fine-resolution pixels with at least 30 soundings. We use two standard regression metrics: normalized RMSE and $R^2$. 

The RMSE is the square root of the mean squared error between the prediction and the true value:

$$RMSE = \sqrt{\frac{\sum_i (y_i - \widehat{y}_i)^2}{N}}$$

where $y_i$ is the true SIF for pixel $i$, $\widehat{y_i}$ is the model's predicted SIF for pixel $i$, and $N$ is the total number for pixels in the evaluation set. In this paper, we further divide RMSE by the average SIF across the train dataset, to get normalized RMSE (NRMSE).

$R^2$ is a measure of how much the variation in the data can be explained by the model predictions. Formally,

$$R^2 = 1 - \frac{\sum_i (y_i - \widehat{y}_i)^2}{\sum_i (y_i - \bar{y})^2}$$

where $\bar{y}$ is the average SIF across the entire test dataset. The top of the fraction is the sum of the squared residuals (difference between true SIF and model prediction). The bottom is the total sum of squares (of the difference between the true SIF and the average SIF across the test dataset), which is proportional to the overall variance of the test data.

\bibliographystyleSM{named}
\bibliographySM{suppref}

\end{document}